%% file: PaperForReview.tex
\documentclass[10pt,twocolumn,letterpaper]{article}

\usepackage{wacv}              

\usepackage{graphicx}
\usepackage{amsmath}
\usepackage{amssymb}
\usepackage{booktabs}

\usepackage{booktabs}
\usepackage{color}
\usepackage{enumitem}
\usepackage{multirow}
\usepackage{algorithm}
\usepackage{algorithmic}
\usepackage{colortbl}
\RequirePackage[format=plain,labelformat=simple,labelsep=period,font=small,compatibility=false]{caption}
\RequirePackage[font=footnotesize,skip=3pt,subrefformat=parens]{subcaption}

\usepackage[pagebackref,breaklinks,colorlinks]{hyperref}

\usepackage[capitalize]{cleveref}
\crefname{section}{Sec.}{Secs.}
\Crefname{section}{Section}{Sections}
\Crefname{table}{Table}{Tables}
\crefname{table}{Tab.}{Tabs.}

\def\wacvPaperID{405} 
\def\confName{WACV}
\def\confYear{2024}

\begin{document}

\title{Controlling Character Motions without Observable Driving Source}

\author{Weiyuan Li\\
Xiaobing.AI\\
{\tt\small liweiyuan@xiaobing.ai}
\and
Bin Dai\\
Xiaobing.AI\\
{\tt\small daibin@xiaobing.ai}
\and
Ziyi Zhou\\
Xiaobing.AI\\
{\tt\small zhouziyi@xiaobing.ai}
\and
Qi Yao\\
Xiaobing.AI\\
{\tt\small yaoqi@xiaobing.ai}
\and
Baoyuan Wang\\
Xiaobing.AI\\
{\tt\small  wangbaoyuan@xiaobing.ai}
}


\twocolumn[{
\renewcommand\twocolumn[1][]{#1}
\maketitle
\begin{center}
  \captionsetup{type=figure}
  \includegraphics[width=1.0\textwidth]{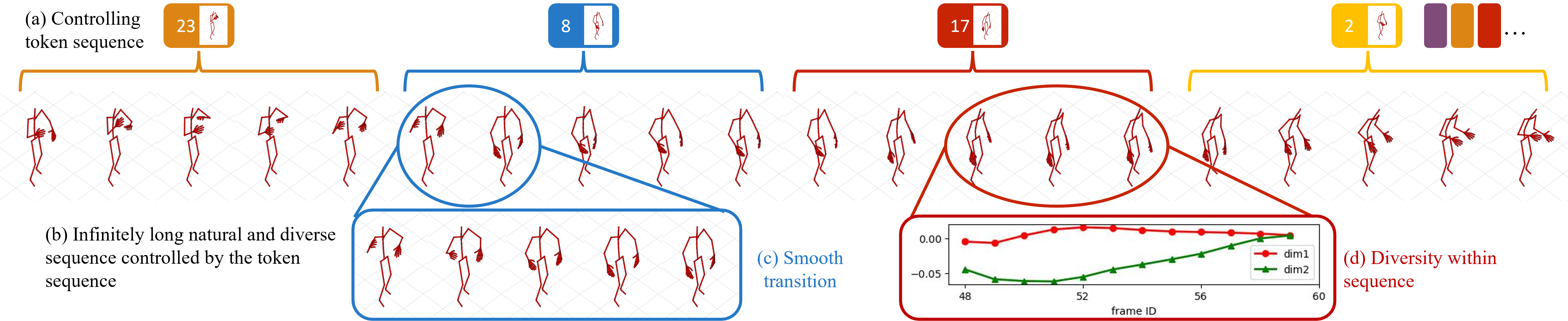}
  \captionof{figure}{We propose an algorithm to generate diverse and unlimited long sequences without a driving source. The sequence is controlled by the tokens produced by a high-level policy. It makes smooth transitions between states and has natural movement within a state.}
  \label{fig:teaser}
\end{center}
}]

\begin{abstract}
   How to generate diverse, life-like, and unlimited long head/body sequences without any driving source? We argue that this under-investigated research problem is non-trivial at all, and has unique technical challenges behind it. Without semantic constraints from the driving sources, using the standard autoregressive model to generate infinitely long sequences would easily result in 1) \textbf{out-of-distribution (OOD) issue} due to the accumulated error, 2) insufficient \textbf{diversity} to produce natural and life-like motion sequences and 3) undesired \textbf{periodic patterns} along the time. To tackle the above challenges, we propose a systematic framework that marries the benefits of VQ-VAE and a novel token-level control policy trained with reinforcement learning using carefully designed reward functions. A high-level prior model can be easily injected on top to generate unlimited long and diverse sequences. Although we focus on no driving sources now, our framework can be generalized for controlled synthesis with explicit driving sources. Through comprehensive evaluations, we conclude that our proposed framework can address all the above-mentioned challenges and outperform other strong baselines very significantly. 
\end{abstract}

\input{inputs/intro}

\begin{figure*}[t!]
    \centering
    \includegraphics[width=0.98\textwidth]{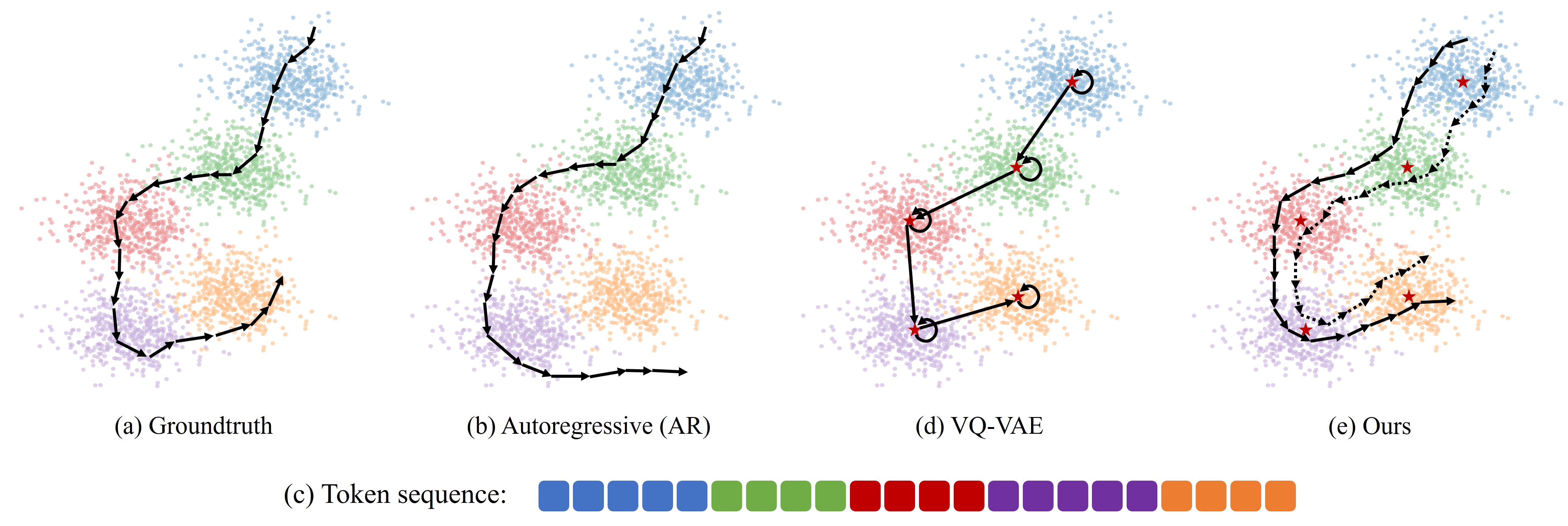}
    \caption{(a) A ground-truth trajectory in the feature space. (b) A trajectory generated by an autoregressive model may suffer from the OOD issue. (c) The token sequence by applying the VQ-VAE encoder and quantizer to the ground-truth trajectory. (d) The trajectory is decoded from the token sequence using the VQ-VAE decoder. (e) The trajectories are decoded from the token sequence using our low-level policy. Our algorithm can produce diverse smooth trajectories.}
    \label{fig:low_level_policy}
\end{figure*}

\input{inputs/related}

\begin{figure*}[t!]
    \centering
    \includegraphics[width=0.96\textwidth]{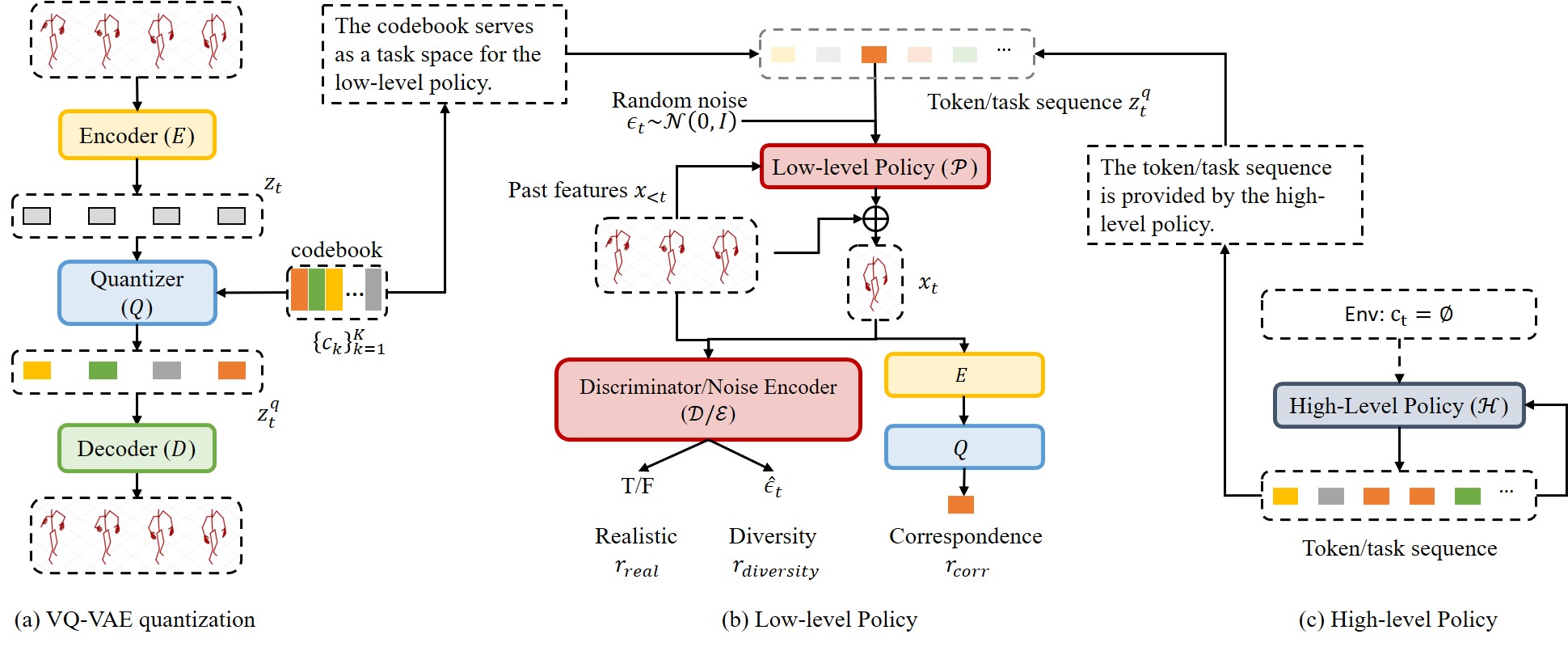}
    \caption{Pipeline. (a) VQ-VAE discretizes the sequence. The token space serves as both the input of the low-level policy and the output of the high-level policy. (b) The low-level policy takes the token, the past features, and a random noise as input and produces the next feature. (c) The high-level policy generates the token sequence.}
    \label{fig:pipeline}
\end{figure*}

\input{inputs/prob_def}

\input{inputs/algo}
\input{inputs/exp}

\section{Conclusion}
\label{sec:conclusion}
We define the problem of generating diverse, life-like, and unlimited long head/body sequences without any driving source. The challenges of the problem are analyzed and a pipeline is proposed to solve this problem. Empirical results show that our algorithm produces significantly better results than previous methods. Moreover, our task space and low-level policy can be re-used for further building more complicated decision modules with multiple driving sources, which will be future work.

{\small
\bibliographystyle{ieee_fullname}
\bibliography{egbib}
}

\end{document}

%% file: inputs/intro.tex
\section{Introduction}
\label{sec:intro}





Recently, synthesizing the character motion (both head and body, rigid or non-rigid) given a certain kind of driving source emerges as a popular topic\cite{zhou2020makelttalk,wang2021audio2head,zhou2021pose,ng2022learning,siyao2022bailando}. For example, Learning2Listen~\cite{ng2022learning} is able to predict both the head pose and non-rigid facial expressions of the listener when given both the audio and video sequence of the speaker. PC-AVS~\cite{zhou2021pose} can generate a sequence of photo-realistic talking heads when given a driving audio, a reference video for the head pose, and a target face image. Similarly, Bailando\cite{siyao2022bailando} can synthesize a dancing body motion sequence for a music audio input. Those models all aim to build \emph{semantic correspondence} between the character motion and the external driving sources, i.e., the lip motion has to be aligned with the corresponding audio segment, and the body motion has to be compatible and harmonized with the rhythm and beat of the driving music. However, there are many scenarios where the character still needs to be continuously animated without an observable driving source, especially when controlling and synthesizing long and versatile character motions that are life-like. For example, in the ``idle" state of a VTuber, when there is no interaction and hence not responding to any external signals, it's indispensable to control the motion behaviors to make it look natural. And perhaps the demand for naturalism is even higher for live streaming of a photo-realistic digital human avatar in order to cross the uncanny valley. Therefore, how to model and control such motion behaviors becomes an urgent yet important research problem, while we move toward the era of the metaverse.

This is not a trivial problem compared to the scenario when the driving source exists. There is one key difference between these two tasks. When the driving source is available, the model is conditioned on the driving information at the same timestamp and thus trained to learn the correspondence between the input and output. However, when the driving source is absent, the only information that the model can condition on is the past sequence, which is also generated by the model itself. We argue that such autoregressive motion synthesis may bring multiple fatal consequences, namely 1) the out-of-distribution issue, 2) the lacking of diversity issue, and 3) the periodic repeating pattern issue. These problems will be carefully analyzed in Sec.~\ref{sec:problem_def}.

To fill the gap and address the above challenges, in this paper, we officially define the problem of controlling character motions without a driving source. Our algorithm consists of three parts: a quantizer that encodes the sequence to a discrete token sequence, a low-level policy to decode the token sequence and a high-level policy to generate the token sequence. Each part is responsible for a challenge mentioned above. 
To tackle the \textbf{out-of-distribution issue}, we discretize the continuous feature space using quantization algorithms like vector-quantization variational autoencoder (VQ-VAE)\cite{van2017neural} to a token sequence. 
Directly decoding the token sequence back into the continuous feature space using the VQ-VAE decoder (as in many previous works\cite{ng2022learning,siyao2022bailando}) will cause the \textbf{lacking of diversity issue}. Therefore, we instead employ a reinforcement learning (RL) framework to replace the VQ-VAE decoder. The policy network, which we call the low-level policy, is responsible for continuously generating the next frame based on the current token and the past frames. In this sense, the token space of the VQ-VAE can also be regarded as a task space for the low-level policy. 
To ensure the policy network capable of generating diverse frames given the same input token, we also add a randomly sampled Gaussian noise as input to the policy network.
Three types of rewards are designed for different purposes: 1) the \emph{realistic reward} forces the generated sequence to look natural; 2) the \emph{diversity reward} is proposed to encourage the policy to produce diverse outputs and 3) the \emph{correspondence reward} requires the produced result to hit the input token.
Finally, to avoid the \textbf{periodic pattern issue} caused by the autoregressive generation procedure, we instead design a random generation scheme, which we call a high-level policy, to produce the token sequence.
It should be noted that, though designed for the scenario without a driving source, such a framework can be generalized for other driving tasks. The discrete token space serves as an interface between the high-level policy and the low-level policy. We only need to change the high-level policy when the driving source exists without modifying the low-level policy. 
To sum up, we make the following contributions:
\vspace{-1mm}
\begin{itemize}[leftmargin=*,itemsep=-3pt]
    \item To our best knowledge, we are the first to define and study the problem of controlling character motion without any driving source. We unveil the problem's importance and its technical challenges.
    \item We design a framework consisting of a high-level policy, a token/task space, and a low-level policy to solve this problem. The token space can also be suited for other high-level policies with driving sources. The low-level policy with carefully designed reward functions can produce natural and diverse results.
    \item We conducted extensive experiments on two public body skeleton datasets and a self-collected VTuber face dataset. Empirical results show that our framework achieves better performance than previous algorithms.
\end{itemize}

%% file: inputs/related.tex
\section{Related work}
\label{sec:related_work}

\paragraph{Face/head/body driving} Prior works mainly focus on driving the face/head/body motion with observable and semantically meaningful sources, which includes speech audio\cite{wang2021audio2head,zhou2020makelttalk,zhou2021pose,suwajanakorn2017synthesizing}, music \cite{siyao2022bailando,huang2020dance,li2022danceformer,chen2021choreomaster}, video\cite{thies2016face2face,hong2022depth,huang2020learning,hsu2022dual,bounareli2022finding} and even text \cite{guo2022generating,zhang2022motiondiffuse,hong2022avatarclip}. 
Among them, the works that aim to accurately control the lip motion to align with the speech audio ~\cite{suwajanakorn2017synthesizing,lahiri2021lipsync3d,thies2020neural,prajwal2020lip,kr2019towards,zhou2019talking} has received much attention. To better control the results, PC-AVS\cite{zhou2021pose} also depends on a separate head pose driving source from a reference video. 
StyleTalker\cite{min2022styletalker} learns an audio-to-motion latent space to produce the head motion. Different from the audio-driven head sequence generation task, Learning2Listen\cite{ng2022learning} tries to generate the head sequence of a listener given both the video and the audio of the speaker. As for the body driving, \cite{alexanderson2020style} drives the body and gesture given the speech audio. Bailando\cite{siyao2022bailando} produces a sequence of dancing skeletons based on the input music. 
Unlike all these tasks which require either one or a few types of observable driving sources, our method aims to generate life-like motion sequences without such driving signals.

\vspace{-3mm}
\paragraph{Reinforcement learning in character control} Physics-based character control task\cite{hodgins1995animating,laszlo1996limit} has a long history. Recently, there are many works trying to solve this problem using reinforcement learning \cite{peng2021acquiring,peng2021amp,peng2022ase,won2022physics}. MotionPrior\cite{peng2021acquiring} regards the controlling problem as a Markov Decision Process (MDP) and trains a policy net to generate the skeleton sequence. It manually defines the state-similarity metric as the reward function. AMP\cite{peng2021amp} replaces the similarity reward function with an adversarial network\cite{goodfellow2020generative} to get rid of the cumbersome human-designed metric. ASE\cite{peng2022ase} further improves diversity by introducing noise into the policy network. Our method is inspired by this line of work. However, Instead of taking a pre-defined goal like \emph{location} or \emph{strike} as input, our policy network takes the token as the task. The reward function corresponding to the task/token is also different. In the pre-defined task case, the reward function is hand designed. However, in our case, we can directly use the VQ-VAE encoder as the reward function. 

\vspace{-3mm}
\paragraph{Quantization + Prior Models.} A two-stage quantization + prior learning model is first proposed in the image synthesis task\cite{van2017neural,razavi2019generating,yu2021vector,esser2021taming}. These algorithms first learn a discrete representation of an image and then train a prior model on this representation. Some works adapted this fashion into text-to-image generation\cite{ramesh2021zero,ramesh2022hierarchical,ding2021cogview,aghajanyan2022cm3}. In these works, a prior condition on the text embedding is learned after obtaining the image discrete representations. This philosophy is also applied to the driving tasks \cite{ng2022learning,siyao2022bailando,hong2022avatarclip}, where a conditional probability distribution is learned on the image tokens conditioned on the driving source. Though our work also uses a similar quantization algorithm as the first step, we have a different purpose for such a design, where we wish to use a discrete space to avoid the OOD issue. 


\vspace{-3mm}
\paragraph{Motion Prediction.} Motion prediction~\cite{ahn2023can,martinez2017human,fragkiadaki2015recurrent,barsoum2018hp,yan2018mt} aims to predict the human motions in the near future based on the past motions. Both deterministic methods~\cite{martinez2017human,fragkiadaki2015recurrent} and probabilistic models~\cite{ahn2023can,barsoum2018hp,yan2018mt} are designed. The key difference between our problem and motion prediction is that we focus on generating unlimited long natural and diverse sequence rather than the near future motion.

%% file: inputs/prob_def.tex
\section{Problem Definition and Analysis}
\label{sec:problem_def}

\paragraph{Problem Definition.} The existing driving problem assumes there is a driving source $c_t$ at each timestep $t$. The driving engine tries to model the probability distribution $p(x_t|c_t)$\footnote{Sometimes there is also an identity input.}, where $x_t$ is the representation of the face/body. 
In this paper, we define the driving problem from a different perspective. Suppose the virtual human is placed in an environment (i.e. live streaming). It should decide its facial expression and body pose at each timestep no matter whether the driving source exists or what the driving source is. Generally speaking, we should model $p(x_t | x_{<t}, c_t)$ for each $t$. This procedure should continuously go on until the whole event ends. 
Unlike previous driving problems that ignores $x_{<t}$ and only models $p(x_t|c_t)$, we consider another scenario when $c_t=\emptyset$ and model $p(x_t|x_{<t})$. Such a setting is even more common in practice since the underlying driving source is often not observable. We argue that this problem is both challenging and important for further building more complicated decision models.

\vspace{-3mm}
\paragraph{Challenges.} 
The first challenge is the OOD issue. Our model $p(x_t|x_{<t})$ is trained on the ground-truth dataset. However, during the inference phase, we apply the model to the dataset generated by the model itself. That being said, the distributions of $x_{<t}$ in the training  and inference phases are different, making the output $x_t$ following an even more different distribution. An illustration is shown in Fig.~\ref{fig:low_level_policy}(a)-(b). The groundtruth trajectory is shown in Fig.~\ref{fig:low_level_policy}(a). The trajectory generated by an autoregressive model is shown in Fig.~\ref{fig:low_level_policy}(b). These two trajectories look similar in the beginning. However, as the number of steps increases, the path becomes significantly different and the generated path may go to the OOD region.

Using VQ-VAE to constrain the output space of the autoregressive model in a discrete inlier set can avoid the OOD issue. Suppose we have trained a VQ-VAE to cluster the dataset into 5 different clusters, each represented in different colors. Fig.~\ref{fig:low_level_policy}(c) shows the corresponding token sequence of the gourndtruth trajectory. However, directly decoding the token sequence will produce a trajectory lacking of diversity, as shown in Fig.~\ref{fig:low_level_policy}(d), which is the second challenge of the problem. A good model should be able to produce diverse and smooth trajectories as shown in Fig.~\ref{fig:low_level_policy}(e).

Lastly, there is also the periodic pattern or the fixed point problem. Though the prior model is conditioned on all the history $x_{<t}$ theoretically, we usually use a fixed length of past window $x_{t-\Delta T:t-1}$ in practice. Suppose we model the distribution $p(x_t|x_{t-\Delta T:t-1})$ as Gaussian and use the mean of the Gaussian distribution as $x_t$ during the inference time, as in many autoregressively regression tasks. Then $x_t$ becomes completely determined by $x_{t-\Delta T:t-1}$. Further, $x_{t+1}$ is then determined by $x_{t-\Delta T +1:t}$, which is again a deterministic function of $x_{t-\Delta T:t-1}$. So we can write $x_{t:t+\Delta T - 1} = f(x_{t-\Delta T:t-1})$, where $f(\cdot)$ is a function correlated to the autoregressive model. The autoregressive model is like applying the same function again and again. Such a process will make the same sequence appear repeatedly. Though we usually add some noise during the sampling procedure in practice, such an issue cannot be completely avoided, as will be presented in our experiments.

%% file: inputs/algo.tex
\section{Method}
\label{sec:algo}

Our method consists of three parts: 1) a discrete token space yielded from a VQ-VAE model, 2) a low-level policy network that decodes the token sequence to the continuous feature space, and 3) a high-level policy that produces the token sequence. The overview pipeline is illustrated in Fig.~\ref{fig:pipeline}. In the following, we will introduce more details for each part.

\subsection{Token Space Derived from VQ-VAE}
\label{sec:vq_vae}

The token space serves as both the output space of the high-level policy and the input space of the low-level policy. 
Once the token space is determined, the high-level policy and the low-level policy can be disentangled and separately developed. The low-level policy only needs to focus on how to generate a natural and diverse sequence based on the provided token sequence, while the high-level policy only cares about how to produce the token sequence based on the environment.
In this paper, we use the quantized latent space of a VQ-VAE model as the token space, which can effectively avoid the out-of-distribution issue since it constrains the output of the high-level policy into a discrete inlier set.

Technically, either a clip-level VQ-VAE or a frame-level VQ-VAE can be adopted. In our current implementation, we adopt a frame-level VQ-VAE. For each frame, the feature $x_t\in\mathbb{R}^d$ is encoded into a continuous latent vector $z_t\in\mathbb{R}^\kappa$ via an encoder $E(\cdot)$, \emph{i.e.} $z_t = E(x_t)$,
where $d$ is the feature space dimension while $\kappa$ is the latent space dimension. The latent vector $z_t$ is then assigned to the nearest code in a learnable codebook $\{c_k\}_{k=1}^K$, where $K$ is the codebook size. Let $Q(\cdot)$ be the quantizer and $q_t$ stand for the code index, then
\begin{equation}
    q_t := Q(z_t) = \mathop{\arg\min}\limits_{k} ||z_t - c_k||_2.
\end{equation}
Denote $z_t^q$ as the $q_t$-th entry in the codebook $\{c_k\}_{k=1}^K$. It is also known as the quantized version of the latent vector $z_t$. The quantized $z_t^q$ is then decoded to the original feature space via a decoder $D(\cdot)$, \emph{i.e.} $\hat{x}_t = D(z_t^q).$

The VQ-VAE, including the encoder, decoder, and codebook, is optimized using the objective
\begin{equation}
    \mathcal{L}_{VQ} = ||x_t - \hat{x}_t||  + ||\text{sg}[z_t] - z_t^q||_2  + \beta ||z_t - \text{sg}[z_t^q]||_2,
\end{equation}
where $\text{sg}[\cdot]$ means stop gradient and $\beta$ is a hyperparameter. Following \cite{van2017neural}, we use $\beta=0.25$ in all our experiments.

\subsection{Low-Level Policy}

Most prior works~\cite{ng2022learning,siyao2022bailando} directly use the VQ-VAE decoder to generate the output given the discrete token sequence. Such a design at least has two drawbacks. Firstly, it may produce unnatural (i.e., flicking) sequences lacking diversity, as shown in Fig.~\ref{fig:low_level_policy}(d). Secondly, it only considers the token at/around the current timestamp when decoding. No long-term dependency of the generated sequence is considered, degrading the generation quality when the sequence becomes infinitely long.

To tackle the issues of the VQ-VAE decoder, we use a reinforcement learning framework, which we call a low-level policy, to decode the token sequence. The state at the current step $t$ includes not only the current token $z_t^q$, but also the past frames $x_{t-\Delta T:t-1}$ for smoothness, and a random sample $\epsilon_t$ for diversity. The policy network $\mathcal{P}(\cdot)$ takes the state $s_t=[z_t^q, x_{t-\Delta T:t-1}, \epsilon_t]$ as input and outputs a $d$-dimensional deviation vector $\delta x_t$ as action. The next frame feature then becomes $x_t=\delta x_t + x_{t-1}$. An illustration is shown in Fig.~\ref{fig:pipeline}(b).

We design three rewards regarding different constraints on the future frame $x_t$. The realistic reward corresponds to whether the sequence $(x_{t-\Delta T:t-1}, x_t)$ looks natural or not. A discriminator is adapted to produce a realistic score. It is an MLP that takes $(x_{t-\Delta T:t-1}, x_t)$ as input. The realistic reward at step $t$ then becomes
\begin{equation}
    r_{t,real} = \mathcal{D}(x_{t-\Delta T:t-1}, x_t).
\end{equation}
The reward is normalized to the range $(0, 1)$ using a sigmoid activation layer. The discriminator is trained in an adversarial manner\cite{goodfellow2020generative}. The loss for the discriminator is
\begin{equation}
    \mathcal{L}_{gan} = \text{CE}\left(\mathcal{D}(x^r_{t^\prime-\Delta T:t^\prime}), 1\right) + \text{CE}\left(\mathcal{D}(x_{t-\Delta T:t}), 0\right),
    \label{eqn:loss_gan}
\end{equation}
where $\text{CE}(\cdot, \cdot)$ stands for the cross entropy function, $x^r$ is a random sequence from the training dataset and $t^\prime$ is a random start frame index.

The second reward regards the correspondence between the output feature $x_t$ and the input token $q_t$. We leverage the trained VQ-VAE encoder $E$ and quantizer $Q$ for this reward. It is designed as
\begin{equation}
    r_{t,corr} = \mathbb{I}(q_t, Q(E(x_t))), 
    \label{eqn:loss_corr}
\end{equation}
where $\mathbb{I}(\cdot,\cdot)$ is an identical function that equals to $1$ when the two arguments are equal and $0$ otherwise.
It requires that $x_t$ should be assigned to $q_t$ by the VQ-VAE encoder and quantizer. However, only using this term will make the training procedure problematic because all the possible $x_t$ are equally bad as long as it does not hit $q_t$. To tackle this issue, we also calculate the change of the $\ell_1$ distance between $x_t$ and $z_t$. This change is further clipped into range $[-1, 0.8]$. If $q_t\neq Q(E(x_t))$, we use the clipped change as the reward, which encourages $x_t$ to move towards $z_t$ when it is assigned to a different token.

The last reward encourages the policy to produce diverse outputs given different noise $\epsilon_t$. Though the policy network takes $\epsilon_t$ as input, it is very likely to completely ignore the noise input without such a diversity reward. To enforce the policy network to encode the information of $\epsilon_t$, we train another noise encoder to reconstruct $\epsilon_t$ given $(x_{t-\Delta T:t-1}, x_t)$. The noise encoder is trained using the L2-Loss
\begin{equation}
    \mathcal{L}_{diverse} = \frac{1}{2} || \epsilon_t - \mathcal{E}(x_{t-\Delta T:t-1}, x_t) ||_2^2,
    \label{eqn:loss_diverse}
\end{equation}
where $\mathcal{E}$ is the noise encoder. In practice, the noise encoder and the discriminator share the same architecture and weights. The diversity reward then becomes
\begin{equation}
    r_{t,diversity} = - || \epsilon_t - \mathcal{E}(x_{t-\Delta T:t-1}, x_t) ||_2^2.
\end{equation}

The final reward function can be written as
\begin{equation}
    r_t = w_r \cdot r_{t,real} + w_c \cdot r_{t,corr} + w_d \cdot r_{t,diversity},
    \label{eqn:reward}
\end{equation}
where $w_r$, $w_c$ and $w_d$ are the weights of each reward. The value at step $t$ is then defined as 
\begin{equation}
    V_t = r_t + \sum_{dt=1}^{+\infty} \gamma^{dt} r_{t+dt},
    \label{eqn:value}
\end{equation}
where $\gamma$ is the discount factor (set as 0.98).

The low-level policy network is trained using proximal policy optimization (PPO)\cite{tang2017exploration}. We use GAE($\lambda$)\cite{schulman2015high} to compute the advantage function and TD($\lambda$)\cite{sutton1998introduction} to update the approximate value function.
The algorithm for training the low-level policy is shown in Algorithm~\ref{alg:low_level}. The discriminator $\mathcal{D}$ and the noise encoder $\mathcal{E}$ are optimized during the training of the policy network.

Note that our low-level policy can also be trained using supervised learning, regarding the reward function (\ref{eqn:reward}) as the loss function. However, using supervised learning will degrade the performance since it ignores the long term dependency. We will empirically demonstrate that using RL will produce better results than supervised learning.

\begin{algorithm}[t!]
\renewcommand{\algorithmicrequire}{\textbf{Input:}}  
\renewcommand{\algorithmicensure}{\textbf{Output:}} 
\caption{Algorithm of Low-Level Policy}
\label{alg:low_level}
    \begin{algorithmic}
        \REQUIRE VQ-VAE encoder $E$, VQ-VAE quantizer $Q$, VQ-VAE codebook, dataset.
        \ENSURE Low-level policy network.
        
        \WHILE{Not converge}
            \STATE Update $\mathcal{D}$ using loss function \ref{eqn:loss_gan}.
            \STATE Update $\mathcal{E}$ using loss function \ref{eqn:loss_diverse}.
            \STATE Update low-level policy network $\mathcal{P}$ using PPO.
        \ENDWHILE
    \end{algorithmic}
\end{algorithm}

\subsection{High-Level Policy}
Generally speaking, the high-level policy, denoted as $\mathcal{H}$, takes both the driving source $c_t$ and the past feature $x_{<t}$ as input and outputs the token for the next step $z_t^q$. So we can write the general form as $z_t^q=\mathcal{H}(x_{<t}, c_t)$. In this paper, we focus on the scenario when $c_t=\emptyset$ and leave the the other types of high-level policies for future work.


A straightforward way to design the high-level policy is to use an autoregressive model that takes the past tokens $z_{<t}^q$ as input and continuously generate the next token, \emph{i.e.} $z_t^q = \mathcal{H}(z_{<t}^q)$. However, as discussed in Section~\ref{sec:problem_def}, using such a model will have the periodic pattern issue. Many previous works manually add some randomness in the sampling procedure to avoid the issue. For example, instead of selecting the token with the highest probability, we can uniformly sample from the top $K$ tokens~\cite{van2017neural}. Considering that our low-level policy can add more diversity to the decoded sequence, we adopt this scheme as one of our high-level policies, denoted as \emph{Ours-A(utoregressive)}.

We also consider using a random prior. Specifically, for each $20$-frame clip, we randomly choose a token from the codebook. With this setting, even the past feature $x_{<t}$ is ignored by the high-level policy. Interestingly, such a simple random strategy can produce the best results in most cases. The generated sequence will first make a natural movement in the region corresponding to the chosen token and then smoothly transit to the next specified region. This scheme is denoted as \emph{Ours-R(andom)}. 

%% file: inputs/exp.tex
\section{Experiments}
\label{sec:exp}

We evaluate our algorithm on two public body datasets, namely the Trinity Gesture dataset~\cite{ferstl2018investigating} and the AIST++ dataset~\cite{li2021learn}, and a self-collected face dataset. The Trinity Gesture dataset includes 224 minutes of body motion and the corresponding audio. The AIST++ dataset contains 1,408 sequences of 3D human dance motion along with the music. In our experiments, we assume that the audio/music is unavailable and only use the body motion data. We also collect 46.4 hours of live-streaming data of a female VTuber. In most of the time, there is indeed no observable driving source but the anchor still have some natural expression and movements. This dataset is split into a training set with 37.4 hours and a test set with 9 hours. Then we detect the face region of the anchor and extract the expression and pose features using EMOCA\cite{danecek2022emoca}. This dataset is named VTuber-EMOCA. We will release the extracted EMOCA features to promote future research works. The implementation details are described in the supplemental material.


\begin{table*}[htbp]
    \centering
    \begin{tabular}{c|cc|cc|cc|cc|cc}
    \hline
    dataset & \multicolumn{4}{c|}{Trinity Gesture} & \multicolumn{2}{c|}{AIST++} & \multicolumn{4}{c}{VTuber EMOCA} \\
    \hline
    \multirow{2}*{Method} & \multicolumn{2}{c|}{Raw space} & \multicolumn{2}{c|}{PCA space} & \multicolumn{2}{c|}{PCA space} & \multicolumn{2}{c|}{Expr.-PCA} & \multicolumn{2}{c}{Pose-PCA} \\
    & FD-5 & FD-10 & FD-5 & FD-10 & FD-5 & FD-10 & FD-5 & FD-10 & FD-5 & FD-10 \\
    \hline 
    GT & 62.86 & 134.1 & 2.25 & 2.18 & 8.42 & 7.75 & 1.68 & 1.68 & 0.51 & 0.55 \\
    Random-$1$ & 240.1 & 565.3 & 12.52 & 9.65 & 78.33 & 17.81 & 24.10 & 7.92 & 7.79 & 1,87 \\
    Random-$5$ & 99.84 & 278.8 & 6.64 & 8.01 & 46.76 & 30.83 & 11.28 & 8.38 & 5.19 & 5.03 \\
    SRandom-$5$ & 76.55 & 185.8 & \underline{2.91} & \underline{2.51} & 19.52 & \underline{10.18} & 5.86 & 6.24 & 1.02 & 0.84 \\
    Autoregressive & 1503 & 3019 & 43.96 & 44.86 & 85.37 & 47.97 & 59.19 & 53.36 & 10.32 & 10.13 \\
    VAE-$5$\cite{kingma2013auto} & 102.0 & 263.9 & 8.79 & 7.85 & 14.21 & 22.58 & 9.32  & 7.09 & 2.09 & 2.32 \\
    VQ-VAE-F\cite{van2017neural} & 136.1 & 280.2 & 8.92 & 7.61 & 22.75 & 21.74 & 4.81 & 3.89 & 2.37 & 2.71 \\
    VQ-VAE-C\cite{ng2022learning} & 130.1 & 262.8 & 9.19 & 7.99 & 32.25 & 33.17 & 16.95 & 16.79 & 2.47 & 2.56 \\
    AMP\cite{peng2021amp} & 143.4 & 294.0 & 8.65 & 7.51 & 22.97 & 19.64 & 9.71 & 8.28 & 3.85 & 3.92 \\
    ASE\cite{peng2022ase} & 81.14 & 171.0 & 3.44 & 3.30 & 13.58 & 12.70 & 1.82 & 2.47 & \textbf{0.42} & 0.70\\
    \hline
    \rowcolor[rgb]{0.902,0.898,0.898} Ours R (No RL) & 71.63 & 155.3 & 4.33 & 5.43 & 12.15 & 11.69 & 3.21 & 4.37 & 0.79 & 0.80 \\
    \rowcolor[rgb]{0.902,0.898,0.898} Ours A & \underline{47.70} & \underline{101.8} & 3.25 & 3.03 & \underline{10.68} & 11.07 & \underline{1.64} & \textbf{1.64} & 0.53 & \underline{0.69}\\
    \hline
    \rowcolor[rgb]{0.902,0.898,0.898} Ours R & \textbf{41.54} & \textbf{92.00} & \textbf{2.15} & \textbf{2.25} & \textbf{9.22} & \textbf{8.89} & \textbf{1.50} & \underline{2.15} & \underline{0.50} & \textbf{0.58}\\
    \hline
    \end{tabular}
    \caption{Quantitative Evaluation on Trinity Gesture, AIST++ and VTuber-Emoca Datasets.}
    \label{tab:fd_comparison}
\end{table*}

\begin{figure*}[t!]
    \centering
    \includegraphics[width=0.95\linewidth]{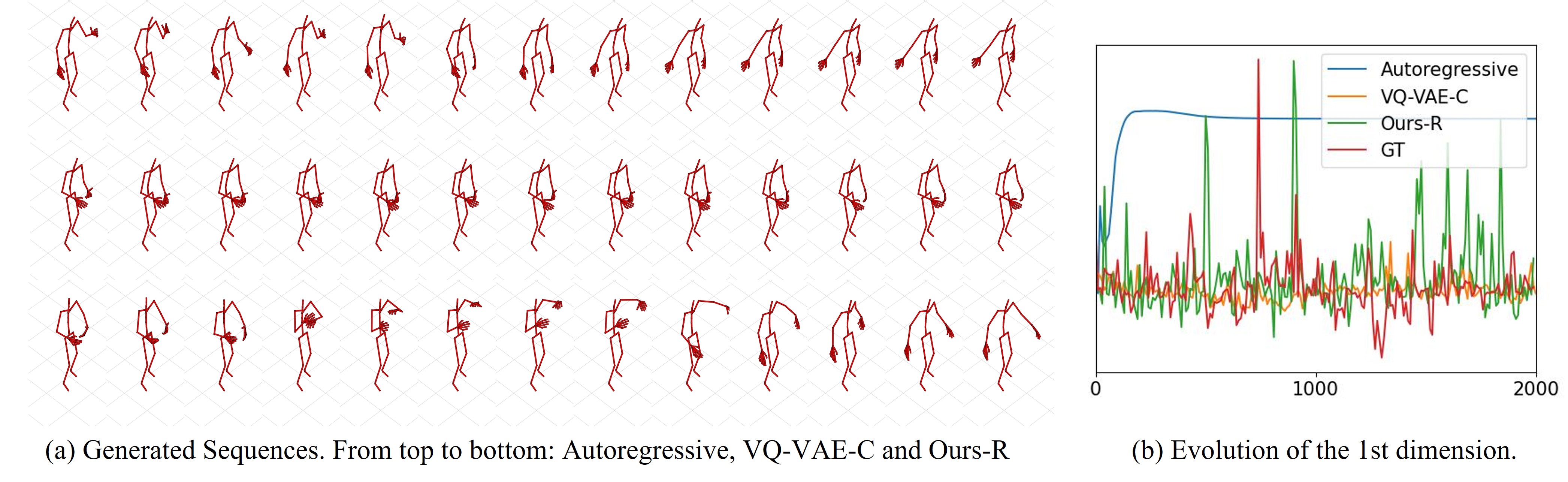}
    \caption{Visualization on Trinity Gesture dataset.}
    \label{fig:gen_seq}
\end{figure*}

\subsection{Quantitative Evaluation}
We first quantitatively compare our algorithm with many previous works that can also be adapted to generate an infinitely long sequence. These baseline methods include

\begin{itemize}[leftmargin=*,itemsep=0pt]
    \item \textbf{Random-$T$: } Randomly select a sequence of length $T$ from the training set at each step and combine all the sequences together as a single long sequence. 
    Though it looks trivial, this is actually a strong baseline because every clip is from the real dataset.
    \item \textbf{SRandom-$T$: } A stronger baseline than \emph{Random-$T$}. We linearly interpolate between every two randomly selected clips in \emph{Random-$T$} to further improve the smoothness. 
    \item \textbf{Autoregressive: } An autoregressive transformer is directly trained on the continuous data using the L2 loss.
    \item \textbf{VAE-$T$: } We train a VAE\cite{kingma2013auto} model on the sequences of length $T$. Then we randomly sample a sequence in the latent space. The sequence is then decoded using the VAE decoder.
    Compared to \emph{Random-$T$}, it does not need to maintain a huge memory pool for the data clips.
    \item \textbf{VQ-VAE\cite{van2017neural}: } Following the quantization + prior model fashion, a VQ-VAE model is first trained on the sequences. Then an autoregressive prior model is further learned on the token sequences. We use both frame-level VQ-VAE as our algorithm does and a clip-level VQ-VAE as Learning2Listen\cite{ng2022learning} does, which are respectively denoted as VQ-VAE-F(rame) and VQ-VAE-C(lip).
    \item \textbf{AMP\cite{peng2021amp}: } Adversarial motion prior (AMP) is also an autoregressive model. It learns a discriminator to distinguish the ground-truth and the generated sequences. The output of the discriminator is used as a reward in the reinforcement learning framework.
    \item \textbf{ASE\cite{peng2022ase}: } Adversarial skill embedding (ASE) further adds noise as input to improve the diversity compared to AMP. Neither AMP nor ASE can use the token sequence to control the output sequence.
\end{itemize}

We evaluate the quality of the generated sequence using the commonly adopted Frechet distance\cite{heusel2017gans} (FD) between the test dataset and the generated sequences. We first generate a set of sequences with 2000 frames\footnote{Generating more frames will produce similar results}. These sequences are then randomly sliced into $T$-frame clips. The mean and variance are computed on the $(T\times d)$-dimensional space. We use FD-$T$ to stand for the FD between the generated sequence and the test sequence with clip length $T$. In our evaluation, we use $T=5$ and $10$.

The performance of different algorithms are shown in Tab.~\ref{tab:fd_comparison}. The best performance is shown in boldface while the second-best method is underlined. Besides comparing baselines, we also provide the ground-truth (GT) performance as a reference by dividing the test set into two groups and calculating the FD between them. We find that the FD distance becomes relatively large in the raw space since the dimension of the raw space is high. To achieve a better sense of these quantitative numbers, we extract the principal components using PCA\footnote{For Trinity Gesture and VTuber-EMOCA expression, we use 20 componets. For the simpler VTuber-EMOCA pose, we use 5 components. For the more complicated AIST++, we use 40 components.} and calculate the FD in the PCA space. For Trinity Gesture dataset, we present FD score in both spaces to show consistency. Ours-R achieves the best performace among all the algorithms in both the raw and the PCA spaces. The autoregressive algorithm completely fails since it drops into a fixed point very quickly as we will see in Sec.~\ref{sec:exp_statistics}. Both VQ-VAE-F and VQ-VAE-C suffer from poor performance since they cannot produce diverse trajectories as demonstrated in Fig.~\ref{fig:low_level_policy}. Replacing the reinforcement learning framework with supervised learning also degrades the performance, though it still outperforms most of the comparing methods.

Considering that the PCA space and the raw space produce consistent rank and the FD scale in the PCA space is kind of normalized, we only report the FD in the PCA space on AIST++ dataset. \emph{Ours-R} again produces the best results among all the algorithms. 
We further validate the performance on our VTuer-EMOCA dataset. Following Learning2Listen~\cite{ng2022learning}, we divide the $56$-dimensional EMOCA feature into a $53$-dimensional expression part and a $3$-dimensional pose part, the results of which are reported separately. Only ASE produces similar results to our methods. Note that ASE does not have a simple interface for high-level policies, yet our algorithm still produces better results on the expression part, which contains most of the dimensions of EMOCA.


\subsection{Qualitative Evaluation}
\label{sec:exp_statistics}

\paragraph{Generation Comparison.} Fig.~\ref{fig:gen_seq}(a) shows the sequence generated by Autoregressive (top), VQ-VAE-C (middle) and Ours-R (bottom) on the Trinity Gesture dataset. The Autoregressive model produces a reasonable sequence in the beginning. However, it gradually moves to the OOD region and stucks in a completely still state. VQ-VAE-C clearly suffers from the lacking of diversity issue. The hands always keep in front of the hip, though every small segment looks natural. Ours-R produces natural and diverse results. We also plot the evolution of the 1st dimension in Fig.~\ref{fig:gen_seq}(b). Ours-R (green curve) has the similar pattern as the groundtruth (red). Autoregressive (blue) soon degenerates to a completely still state while VQ-VAE-C (orange) has relatively small amplitude.

\vspace{-0.3cm}
\paragraph{Controllable Generation.} Fig.~\ref{fig:teaser} shows how we can control the generation by specifying the token sequence. The top row is the high-level token sequence\footnote{The token sequence can be specified by any kind of high-level policy. In our case, we generate the token sequence using Ours-R.}. For the first four tokens, we visualize the direct decoding via the VQ-VAE decoder. The corresponding task can be interpreted as \emph{``generating a natural sequence around the direct decoding"}. The middle row shows the frames generated by our low-level policy based on the provided token sequence. Each token corresponds to 20 frames and we display one of every 4 frames. Fig.~\ref{fig:teaser}(c) shows the details from the 21st to the 25th frames. We can see that our algorithm produces smooth transitions between different tokens. Fig.~\ref{fig:teaser}(d) plots the evolution of the first two dimensions from the 48th to 59th frames. These frames are all similar to the direct decoding. However, they do not keep completely still. Rather, they have a natural tiny movement around the specified token.

\begin{figure}
    \centering
    \includegraphics[width=0.48\linewidth]{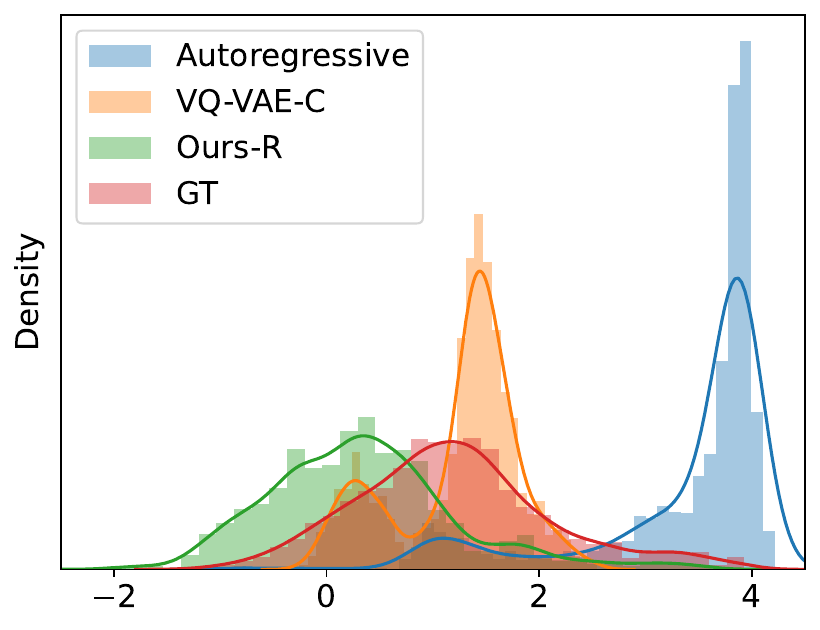}
    \includegraphics[width=0.48\linewidth]{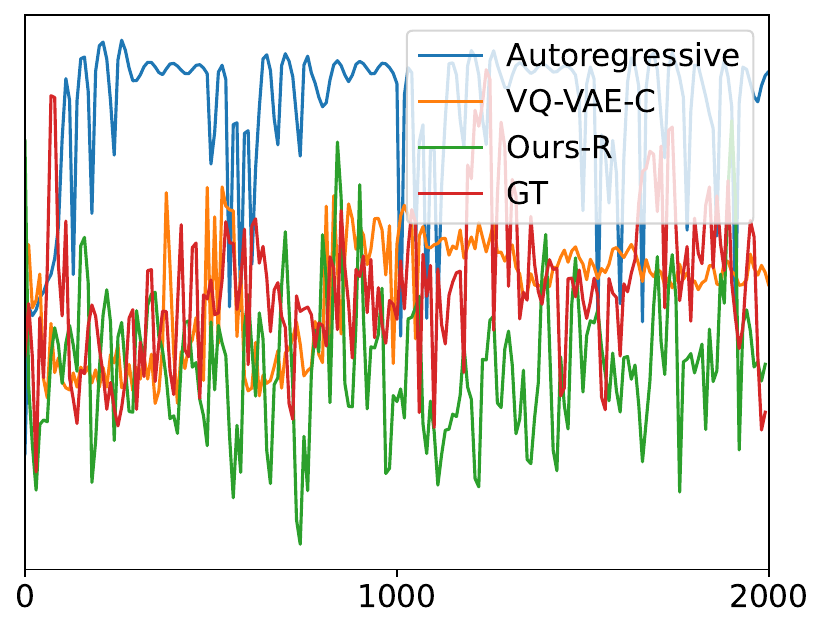}
    \caption{(left) Distribution and (right) evolution of the first dimension of the generated VTuber-EMOCA sequences.}
    \label{fig:gen_seq_dist}
\end{figure}

\vspace{-0.3cm}
\paragraph{Addressing Challenges.} In Sec.~\ref{sec:problem_def}, we analyzed three challenges in this problem. We then empirically address all the challenges. We visualize the distribution of the first dimension of the generated sequences on VTuber-EMOCA dataset in Fig.~\ref{fig:gen_seq_dist}(a). Autoregressive clearly has the OOD issue. Using a discrete space can effectively avoid this issue. Both VQ-VAE-C and our algorithm range in the region where the groundtruth has high density. However, the distribution of VQ-VAE-C is more concentrated, meaning that it suffers from the lacking of diversity issue. We also plot the evolution of the first dimension in Fig.~\ref{fig:gen_seq_dist}(b). The autoregressive algorithm generates a periodic pattern though the pattern in each period is slightly different. The periodic pattern issue is not that obvious in the VQ-VAE-C algorithm because there are also some additional randomness during sampling. However, we do observe that some patterns repeatedly appear in the token sequence.

\subsection{Ablation Study}
We designed three different rewards for different purposes in the low-level policy framework. In this subsection, we apply an ablation study to demonstrate how each reward contributes to the whole framework. Besides the quality metrics FD-5 and FD-10, we use another two metrics regarding correspondence and diversity. We use the hit rate to evaluate the correspondence between the generated sequence and the controlling task token. It counts how many percentages of the generated frames actually hits the task token. To evaluate the diversity, we first cluster the frames in the training data into $100$ clusters. During generation, we generate multiple sequences for each initialization. The entropy of the cluster-ID histogram is calculated for each initialization. We report the average entropy over different initialization as \emph{Div}. The result on the Trinity Gesture dataset is shown in Tab.~\ref{tab:ablation_trinity}.

Using all the three types of rewards produces the best FD score. Removing the discriminator (the second last row) significantly degrades the performance. Comparing the Div between the first section and the second section indicates that $r_{diversity}$ indeed benefits improving the diversity of the generated sequence. Lastly, we will not be able to control the low-level policy if we remove the $r_{corr}$. Though the hit rate seems not that high, we explain it as a necessary sacrifice for generating smooth sequences. 

\begin{table}[t!]
    \centering
    \begin{tabular}{ccc|cccc}
        \hline
        Dis. & Cor. & Noi. & FD-5$\downarrow$ & FD-10$\downarrow$ & Hit$\uparrow$ & Div$\uparrow$ \\
        \hline 
        \checkmark &  &  & 8.65 & 7.51 & -- & 2.67  \\
         & \checkmark &  & 6.88 & 5.78 & 0.16 & 4.14  \\
        \checkmark & \checkmark &  & 5.31 & 5.11 & 0.05 & 4.27  \\
        \hline 
        \checkmark &  & \checkmark & 3.44 & 3.30 & -- & 5.20 \\
         & \checkmark & \checkmark & 8.15 & 6.36 & 0.19 & 4.27 \\
        \checkmark & \checkmark & \checkmark & 2.15 & 2.25 & 0.14 & 5.01 \\
        \hline 
    \end{tabular}
    \caption{Ablation Study on Trinity Gesture Dataset.}
    \label{tab:ablation_trinity}
\end{table}